%% file: 00_main.tex
\documentclass{article}
\usepackage{ijcai24}

\usepackage{times}
\usepackage{soul}
\usepackage{url}
\usepackage[hidelinks]{hyperref}
\usepackage[utf8]{inputenc}
\usepackage[small]{caption}
\usepackage{graphicx}
\usepackage{amsmath}
\usepackage{amsthm}
\usepackage{booktabs}
\usepackage{algorithm}
\usepackage{algorithmic}
\usepackage[switch]{lineno}
\usepackage{xcolor}

\title{WWW: Where, Which and Whatever Enhancing Interpretability in Multimodal Deepfake Detection}

\author{
Juho Jung$^1$\thanks{These authors contributed equally to this work.}\and
Sangyoun Lee$^2$\footnotemark[1]\and
Jooeon Kang$^2$\footnotemark[1]\And
Yunjin Na$^3$\footnotemark[1]\\
\affiliations
$^1$Sungkyunkwan University\and
$^2$Sogang University\and
$^3$Seoul National University
\emails
jhjeon9@g.skku.edu\and
leesy0882@sogang.ac.kr\and
jekang@sogang.ac.kr\and
lumierej@snu.ac.kr
}
\begin{document}

\maketitle

\begin{abstract}
    All current benchmarks for multimodal deepfake detection manipulate entire frames using various generation techniques, resulting in oversaturated detection accuracies exceeding 94\% at the video-level classification. However, these benchmarks struggle to detect dynamic deepfake attacks with challenging frame-by-frame alterations presented in real-world scenarios. To address this limitation, we introduce \emph{\textbf{FakeMix}}, a novel clip-level evaluation benchmark aimed at identifying manipulated segments within both video and audio, providing insight into the origins of deepfakes. Furthermore, we propose novel evaluation metrics, \emph{\textbf{Temporal Accuracy (TA)}} and \emph{\textbf{Frame-wise Discrimination Metric (FDM)}}, to assess the robustness of deepfake detection models. Evaluating state-of-the-art models against diverse deepfake benchmarks, particularly \emph{\textbf{FakeMix}}, demonstrates the effectiveness of our approach comprehensively. Specifically, while achieving an Average Precision (AP) of 94.2\% at the video-level, the evaluation of the existing models at the clip-level using the proposed metrics, TA and FDM, yielded sharp declines in accuracy to 53.1\%, and 52.1\%, respectively. Code is available at \texttt{\href{https://github.com/lsy0882/FakeMix}{https://github.com/lsy0882/FakeMix}}.
\end{abstract}

\input{01_Introduction}
\input{02_Relatedwork}
\input{03_Method}
\input{04_Experiment}
\input{05_Conclusion}

\appendix
\section*{Ethical Statement}
There are no ethical issues.

\section*{Acknowledgments}
This research was supported by the Institute of Information \& communications Technology Planning \& Evaluation (IITP) grant funded by Korea government (MSIT) (RS-2022-00143911, AI Excellence Global Innovative Leader Education Program).

\bibliographystyle{named}
\bibliography{ijcai24}
\end{document}

%% file: 01_Introduction.tex
\section{Introduction}

Rapid advances in hyper-realistic deepfake technology~\cite{zhang2022deepfake,seow2022comprehensive,guarnera2020deepfake,singh2020using,korshunova2017fast} have raised significant privacy and social concerns~\cite{chen2022towards,li2021detecting}, requiring robust detection methods across both video \cite{tolosana2020deepfakes} and audio domains \cite{jia2018transfer}. Despite progress in multimodal deepfake detection, existing benchmarks such as FakeAVCeleb~\cite{khalid2021fakeavceleb}, DFDC~\cite{dolhansky2020deepfake} and KoDF~\cite{kwon2021kodf}, focus on full-video manipulations, which often results in inflated detection accuracy and a lack of insight into specific manipulated segments. This gap highlights the need for more precise detection methodologies that can identify and analyze the manipulated regions within the media.

\begin{figure}[t]
  \centering
  \includegraphics[width=\columnwidth]{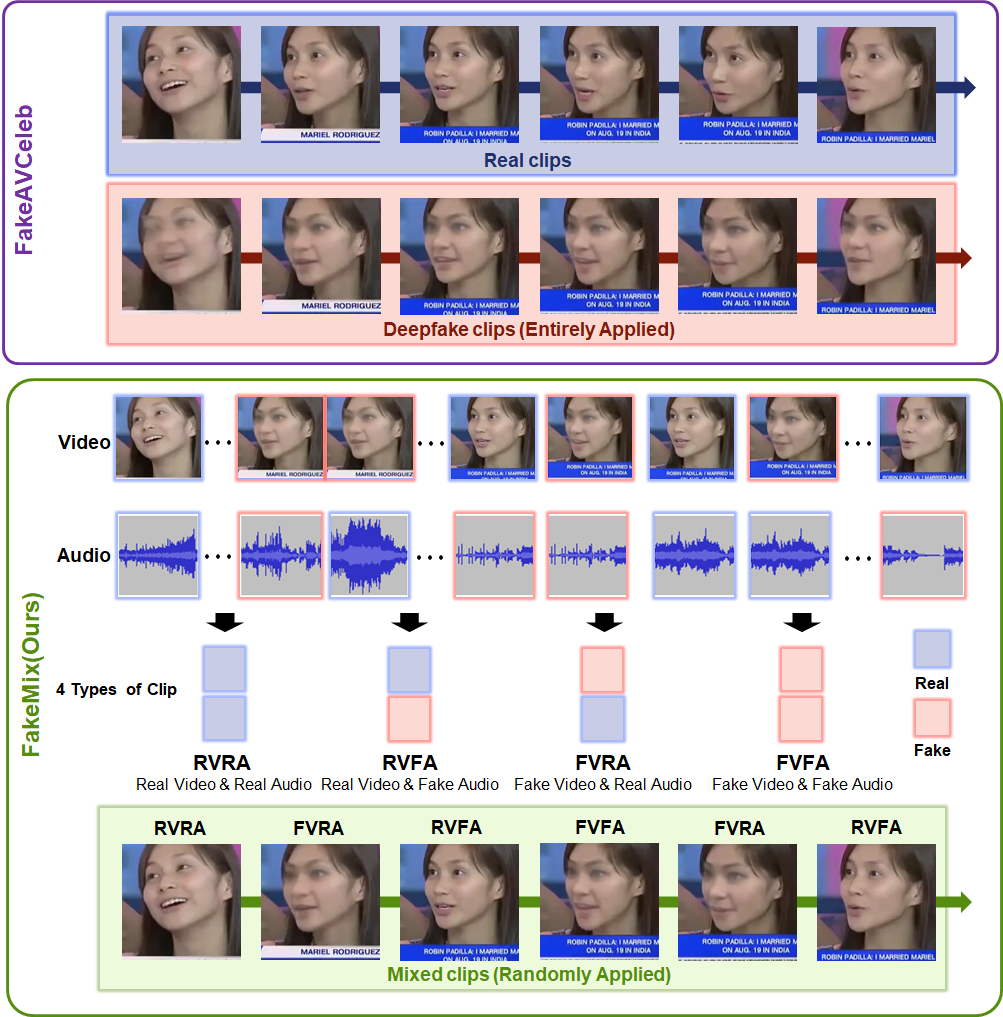}
  \caption{Comparison between previous benchmark and the proposed benchmark, \emph{FakeMix}. While \emph{FakeAVCeleb} operated deepfake on complete video or audio segments for video-level classification, \emph{FakeMix} introduces dynamic frame-level alterations to enhance evaluation of deepfake video detection.}
  \label{fig:prev_bench_vs_fakemix}
\end{figure}

\begin{table*}[t]
\centering
\caption{Comparison of Benchmark Datasets for Deepfake Detection}
\small
\begin{tabular}{c|c|c|c|c}
\hline
\textbf{Dataset} & \textbf{Fake Video} & \textbf{Fake Audio} & \textbf{Fine-grained labeling} & \textbf{Deepfake Appliance} \\ \hline
DFDC \cite{dolhansky2020deepfake} & Yes & Yes & No & Entirely applied\\ \hline
KoDF \cite{kwon2021kodf} & Yes & No & No & Entirely applied\\ \hline
FakeAVCeleb \cite{khalid2021fakeavceleb} & Yes & Yes & Yes & Entirely applied\\ \hline
\textbf{FakeMix (Ours)} & Yes & Yes & Yes & Randomly applied to specific segments \\ \hline
\end{tabular}
\end{table*}

Recent works have evolved from image-based methods~\cite{khan2021video,tarasiou2020extracting,zheng2021exploring,zhang2021detecting,gu2022region,gu2022hierarchical,heo2023deepfake}, which analyze facial information~\cite{ikram2023performance,heo2023deepfake} or morphological details~\cite{tarasiou2020extracting,li2021deepfake}, to more sophisticated video-based methods~\cite{yang2023avoid,wang2022m2tr,lewis2020deepfake,khalid2021fakeavceleb,shahzad2022lip,hashmi2022multimodal,cai2022you,khalid2021evaluation} that incorporate temporal data~\cite{gu2022hierarchical}. However, these approaches generally rely on binary classification of the entire videos (video-level), overlooking specific regions of manipulation and thereby limiting a comprehensive assessment of detection model performance. Moreover, as deepfakes become more sophisticated, the importance of temporal information in identifying inconsistencies in facial movements \cite{jung2023safe} has become more pronounced, highlighting the shortcomings of current methodologies. Recognizing these challenges, our work introduces three main contributions as follows: 

\begin{enumerate}
    \item We propose \emph{FakeMix}, a novel clip-level audio-video multimodal deepfake detection benchmark. Unlike established benchmarks that dominantly focus on overall video-level manipulation, \emph{Fakemix} provides a distinctive assessment by pinpointing specific tampered segments within contents. This approach addresses critical limitations of current benchmarks, which overlook localized alterations.
    \item We develop novel evaluation metrics, namely Temporal Accuracy (\emph{TA}) and Frame-wise Discrimination Metric (\emph{FDM}), designed to validate the robustness of deepfake detection models. These metrics enable precise identification of deepfake-affected regions, enhancing the granularity of results. Our comprehensive evaluation against existing benchmarks demonstrates the efficacy and necessity of incorporating \emph{TA} and \emph{FDM} into the evaluation framework.
    \item To the best of our knowledge, this is the first attempt to assess deepfake video detection at the clip-level, aiming to enhance interpretability. By precisely identifying the specific location (\textbf{W}here), modality (\textbf{W}hich), and deepfake generation technique (in \textbf{W}hatever benchmarks) employed in the manipulation, our approach represents a significant rectification, offering insights into understanding multimodal deepfake detection.
\end{enumerate}

%% file: 02_Relatedwork.tex
\section{Related Work}
There have been numerous research works that studied how to detect deepfakes in multimedia. Recently, deepfake detection studies leverage various DNN architectures to identify and distinguish manipulated videos~\cite{SurveyDLforDF2022,yu2021survey}.
Depending on which modalities are involved, deepfake detection tasks can be divided as follows:

\textbf{Single-Modality Deepfake Video Detection.}
In general, conventional methods utilized a single modality, especially visual domain. \cite{SharpMultiInstance2020} addressed the challenge of partial face manipulations, where only video-level labels are provided.
\cite{SpatioTemporalInconsistency2021} exploited spatial-temporal inconsistency appeared in forged videos. To tackle the poor generalization issue, \cite{IdReveal2021} enhanced robustness through metric learning with adversarial training to capture temporal facial features, which incorporates high-level semantic features.
In spite of their effectiveness, they do not guarantee the high performance of videos with audio deepfakes all at once. This demonstrates the need for a methodology that utilizes both modalities simultaneously.

\textbf{Audio-visual Deepfake Video Detection.}
The emergence of multimodal learning has led to the development of deepfake detection works integrating both auditory and visual modalities.
\cite{zhou2021joint} presented a task for joint audio-video deepfake detection, leveraging intrinsic synchronization between modalities. They improved generalization abilities in unseen deepfake types, focusing on modality relationships.
To this end, \cite{zhao2022self} introduced a self-supervised transformer-based contrastive learning. They leveraged learning lip motion without extensive annotations, encouraging alignment of paired audio-visual representations while promoting diversity on unpaired instances.
\cite{feng2023self} developed an auto-regressive model to generate audio-visual feature sequences, capturing temporal synchronization. 
\cite{yu2023modality} introduced a unified modality-agnostic approach to handle missing modality scenarios and extract speech correlation, making deepfakes challenging to reproduce.
\cite{raza2023multimodaltrace} also proposed a unified framework, which extracts and fuses learned channels from audio and video for effective multi-label detection.
While these studies have exploited significant techniques to improve detection performance, they can only detect whether deepfakes are occurred in the entire video or audio unit.

\textbf{Existing Benchmarks of Deepfake Detection.}
The evolution of deepfake detection has been highlighted by key benchmarks, summarized in Table 1, including DFDC~\cite{dolhansky2020deepfake}, KoDF~\cite{kwon2021kodf}, and FakeAVCeleb~\cite{khalid2021fakeavceleb}.
However, in response to the increasing complexity of deepfake techniques, they are showing obvious limitations.
Existing benchmarks only involve scenarios where deepfakes are applied to every frame within the video, which cannot fully represent various real-world applications, where deepfakes can be applied to specific segments of the video. They also lack attention to delicate manipulations, such as minor changes in facial expressions or specific features.
Although KoDF and FakeAVCeleb have attempted to incorporate culturally specific representations and audio-visual elements, the problem of detecting partial deepfakes remains unresolved.

%% file: 03_Method.tex
\section{Methodology}

\begin{table*}[ht]
\centering
\caption{Comparison of the performance of deepfake detection models on the established deepfake benchmark, FakeAVCeleb and the proposed benchmark, \emph{FakeMix}.}
\label{tab:Comparison of the same model's performance on existing benchmarks and the FakeMix benchmark}
\begin{tabular}{ccccccc}
\toprule
\textbf{Benchmark} & \textbf{Model} & \textbf{Modality} & \textbf{Task} & \textbf{Acc} & \textbf{TA} & \textbf{FDM} \\
\midrule
FakeAVCeleb & Xception \cite{khalid2021fakeavceleb} & A & video-level & 0.7306 & - & - \\
\textbf{FakeMix} & Xception \cite{khalid2021fakeavceleb} & A & clip-level & - & 0.5905 & 0.6018  \\
FakeAVCeleb & Xception \cite{khalid2021fakeavceleb} & V & video-level & 0.7626 & - & - \\
\textbf{FakeMix} & Xception \cite{khalid2021fakeavceleb} & V & clip-level & - & 0.5060 & 0.5034  \\
\midrule
FakeAVCeleb & AVAD \cite{feng2023self} & A-V & video-level & 0.9420 & - & - \\
\textbf{FakeMix} & AVAD \cite{feng2023self} & A-V & clip-level & - & 0.5312 & 0.5212 \\
\bottomrule
\end{tabular}
\end{table*}

\subsection{FakeMix}
To mitigate constraints in existing benchmarks, our work introduces a new benchmark \emph{FakeMix}, a novel clip-level assessment technique for evaluating the robustness and generalization of multimodal deepfake detection. \emph{FakeMix} is designed to address sophisticated scenarios where deepfakes are randomly applied to specific segments of the video and audio, offering more realistic conditions and emphasizing multimodal alignment to enhance interpretability of deepfake detection models. Unlike previous benchmarks, as depicted in Figure~\ref{fig:prev_bench_vs_fakemix}, \emph{FakeMix} incorporates random segment insertions in clips by manipulating both video and audio within one-second intervals to measure the adaptability of multimodal deepfake detection models. 

\subsection{Generation and Description of FakeMix}
As shown in Figure~\ref{fig:prev_bench_vs_fakemix}, let \( V_r = \{v_{r1}, v_{r2}, \dots, v_{rn}\} \) and \( V_f = \{v_{f1}, v_{f2}, \dots, v_{fm}\} \) represent the sets of clips from Real Video and Fake Video, respectively, where \( v_{ri} \) and \( v_{fi} \) denote the \( i \)-th clip in each set. Similarly, let \( A_r = \{a_{r1}, a_{r2}, \dots, a_{rn}\} \) and \( A_f = \{a_{f1}, a_{f2}, \dots, a_{fm}\} \) represent the sets of clips from Real Audio and Fake Audio, respectively, where \( a_{ri} \) and \( a_{fi} \) represent the \( i \)-th clip in each set. To create a \emph{FakeMix} video sequence \( V \), we randomly select clips from either \( V_r \) or \( V_f \) and concatenate them. Similarly, to create a \emph{FakeMix} audio sequence \( A \), we randomly select clips from either \( A_r \) or \( A_f \) and concatenate them as follows: 

\begin{itemize}
    \item Randomly selecting clips for the video sequence:
    \begin{equation}
    V = \{ v_{ij} \mid v_{ij} \in V_r \cup V_f \}
    \end{equation}
    \item Randomly selecting clips for the audio sequence:
    \begin{equation}
    A = \{ a_{ij} \mid a_{ij} \in A_r \cup A_f \}
    \end{equation}
\end{itemize}
Here, \( i \) and \( j \) denote the indices of the selected clips from the respective sets. 

Consequently, within \emph{FakeMix}, the videos are categorized at the clip-level as Real or Fake for both video and audio. Hence, the generated videos by \emph{FakeMix} can be utilized to determine the segments within a video where deepfake manipulation occurs in both video and audio components.

\subsection{Evaluation Metrics for FakeMix}
As illustrated in Figure 2, we employ two novel metrics designed to offer a more granular analysis of deepfake detection capabilities: \emph{Temporal Accuracy (TA)} and \emph{Frame-wise Discrimination Metric (FDM)}. These metrics allow us to assess the effectiveness of deepfake detection at the individual frame level, which is critical for identifying and comprehending the temporal dynamics of deepfake manipulations.

\begin{figure}[h]
  \centering
  \includegraphics[width=\columnwidth]{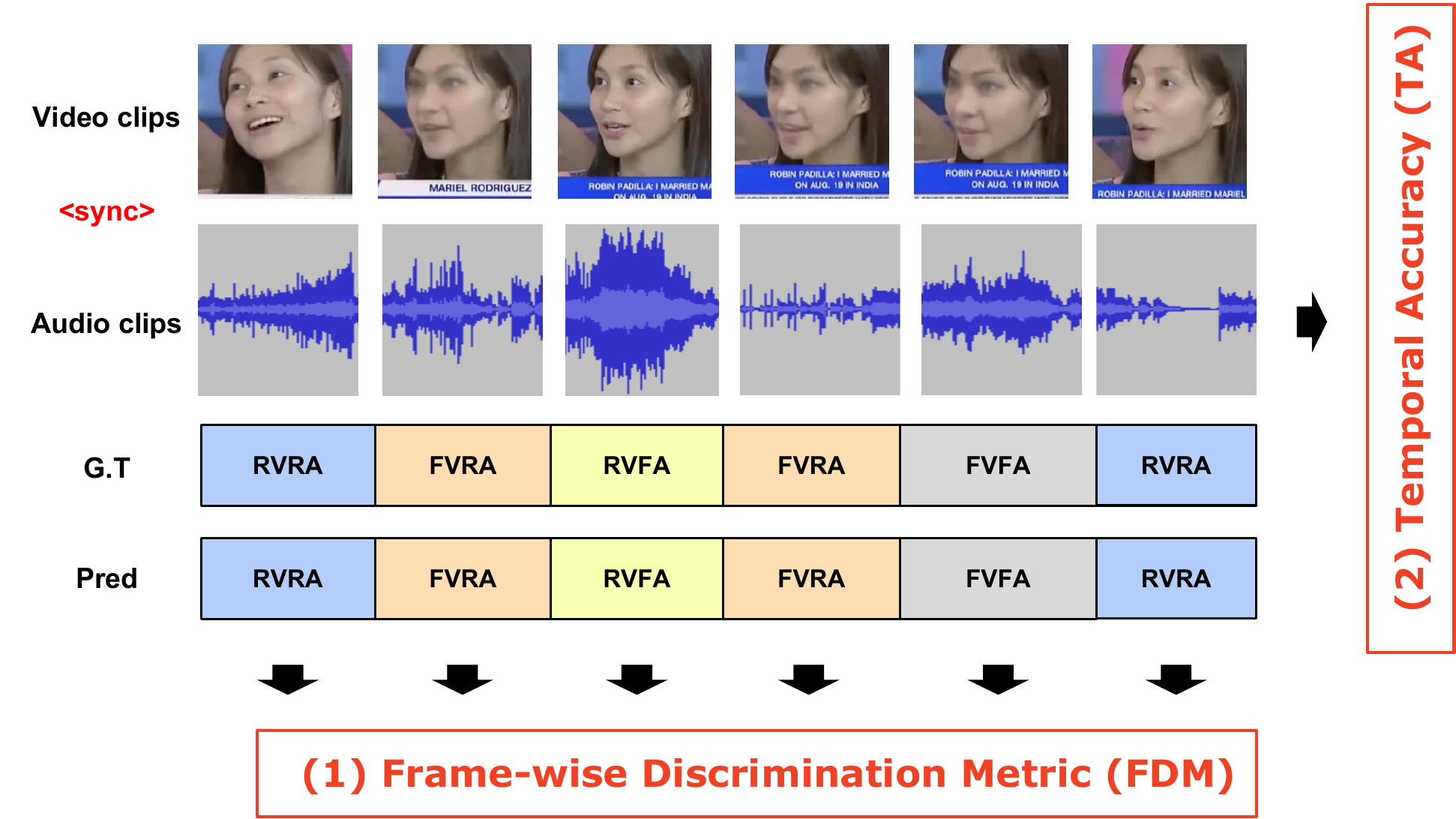}
  \caption{Comprehensive overview of \emph{Temporal Accuracy} and \emph{Frame-wise Discrimination Metric} conducted on the \emph{Fakemix}.}
  \label{fig:eval_metrics}
\end{figure}

\subsubsection{Temporal Accuracy (TA)}
TA is a metric utilized to gauge the frame-level precision of deepfake detection models in predicting the authenticity of each frame within a video. This metric is defined as:

\begin{equation}
    \text{TA} = \frac{1}{N} \sum_{i=1}^{N} \left( \frac{1}{F_{v_i}} \sum_{j=1}^{F_{v_i}} I(\hat{y}_{ij} = y_{ij}) \right),
\end{equation}
where $N$ is the number of videos, $F_{v_i}$ is the total number of frames in the $i$-th video, $\hat{y}_{ij}$ is the predicted label for the $j$-th frame of the $i$-th video, $y_{ij}$ is the ground truth, and $I$ is the indicator function which is 1 if the predicted label equals the ground truth and 0 otherwise.

\subsubsection{Frame-wise Discrimination Metric (FDM)}
To complement TA, we introduce the FDM, which assesses the model's discrimination accuracy over the entire dataset at the frame level. It is expressed as:

\begin{equation}
    \text{FDM} = \frac{\sum_{i=1}^{N} \sum_{j=1}^{F_{v_i}} I(\hat{y}_{ij} = y_{ij})}{\sum_{i=1}^{N} F_{v_i}},
\end{equation}

In this equation, $N$ represents the number of videos, and $F_{v_i}$ signifies the count of frames within the $i$-th video. Here, $I(\hat{y}_{ij} = y_{ij})$ computes the correctness of predictions at the individual frame level.

These metrics are pivotal as they provide a detailed understanding of a model's ability to discern real and fake content at a granular level, echoing the need for sophisticated evaluation in the age of advanced deepfakes. With TA and FDM, we aim to establish a standard that can effectively measure and guide the development of next-generation deepfake detection models.

%% file: 04_Experiment.tex
\section{Experiments}
The difference in evaluation results between FakeAVCeleb, which evaluates the models at the video level, and FakeMix, which evaluates at the clip level, demonstrates that FakeMix is more suitable for assessing the robustness and generalization of deepfake detection models.

\subsection{Experimental Settings}

As shown in Table~\ref{tab:Comparison of the same model's performance on existing benchmarks and the FakeMix benchmark}, we first conducted an experiment to identify the differences in evaluation methodologies within a single modality by assessing the same model across each dataset. During this experiment, we utilized the Xception model previously used by \cite{khalid2021fakeavceleb}, and maintained consistent data preprocessing and hyperparameter settings.

Subsequently, we evaluated the robustness of the AVAD model, as proposed by \cite{feng2023self}, across different modalities by testing it on each dataset. Data preprocessing for the FakeAVCeleb dataset, which contains longer video sequences, involved using sequences of length $N = 50$ from 2.0-second videos. In contrast, for the shorter 1-second clips of the FakeMix dataset, we adjusted the sequence length to $N = 50$ from 1.0-second videos. To address the reduced video duration, we scaled the probability scores output by the AVAD model, considering scores of 0.5 or higher as indicative of fakes. This scaling was critical for accurate computation of True Acceptance (TA) and False Detection Metrics (FDM). Aside from these dataset-specific preprocessing modifications, all other experimental settings conformed to those outlined by \cite{feng2023self}.

\subsection{Results}
In the FakeAVCeleb benchmark, which tests video-level classification, the Xception model achieves approximately 76\% accuracy. Meanwhile, another model, AVAD, reaches a higher accuracy of 94\% in the same video-level classification on FakeAVCeleb. However, in the FakeMix benchmark, which tests clip-level classification, both Xception and AVAD models show a reduction in accuracy to around 50-60\%. Notably, while the Xception model demonstrated lower performance than AVAD at the video level, it outperforms AVAD in clip-level performance.

The experimental results indicate that video-level classification often leads to overestimation, as the entire video is labeled as deepfake even if only a portion of the video contains manipulated content. Therefore, evaluating deepfake detection models at the clip or frame level is crucial to accurately verify their effectiveness. Our proposed evaluation metrics offer a more significant understanding and interpretability in deepfake detection, enabling precise identification of manipulated segments within contents. This approach enhances the reliability and applicability of deepfake detection models in practice.

%% file: 05_Conclusion.tex
\section{Conclusion}
The surge in hyper-realistic deepfake techniques has raised concerns about the authenticity of video and audio content. However, existing multimodal deepfake benchmarks often overlook specific manipulated segments, resulting in inflated detection accuracy and a lack of insight. To address this exaggerated efficacy, we introduced \emph{FakeMix}, a clip-level evaluation benchmark that enhances interpretability by targeting manipulated video-audio segments. Additionally, our proposed evaluation metrics, \emph{TA} and \emph{FDM}, effectively assess the robustness and reliability of deepfake detection methods. By rethinking the overall assessment framework, these findings which highlight the importance of adopting clip-level assessments and refined evaluation metrics, lay the groundwork for more comprehensive and accurate deepfake detection strategies to combat deceptive content.

%% file: 00_main.bbl
\begin{thebibliography}{}

\bibitem[\protect\citeauthoryear{Cai \bgroup \em et al.\egroup }{2022}]{cai2022you}
Zhixi Cai, Kalin Stefanov, Abhinav Dhall, and Munawar Hayat.
\newblock Do you really mean that? content driven audio-visual deepfake dataset and multimodal method for temporal forgery localization.
\newblock In {\em 2022 International Conference on Digital Image Computing: Techniques and Applications (DICTA)}, pages 1--10. IEEE, 2022.

\bibitem[\protect\citeauthoryear{Chen \bgroup \em et al.\egroup }{2022}]{chen2022towards}
Zhaoyu Chen, Bo~Li, Jianghe Xu, Shuang Wu, Shouhong Ding, and Wenqiang Zhang.
\newblock Towards practical certifiable patch defense with vision transformer.
\newblock In {\em Proceedings of the IEEE/CVF Conference on Computer Vision and Pattern Recognition}, pages 15148--15158, 2022.

\bibitem[\protect\citeauthoryear{Cozzolino \bgroup \em et al.\egroup }{2021}]{IdReveal2021}
Davide Cozzolino, Andreas R{\"o}ssler, Justus Thies, Matthias Nie{\ss}ner, and Luisa Verdoliva.
\newblock Id-reveal: Identity-aware deepfake video detection.
\newblock In {\em Proceedings of the IEEE/CVF International Conference on Computer Vision}, pages 15108--15117, 2021.

\bibitem[\protect\citeauthoryear{Dolhansky \bgroup \em et al.\egroup }{2020}]{dolhansky2020deepfake}
Brian Dolhansky, Joanna Bitton, Ben Pflaum, Jikuo Lu, Russ Howes, Menglin Wang, and Cristian~Canton Ferrer.
\newblock The deepfake detection challenge (dfdc) dataset.
\newblock {\em arXiv preprint arXiv:2006.07397}, 2020.

\bibitem[\protect\citeauthoryear{Feng \bgroup \em et al.\egroup }{2023}]{feng2023self}
Chao Feng, Ziyang Chen, and Andrew Owens.
\newblock Self-supervised video forensics by audio-visual anomaly detection.
\newblock In {\em Proceedings of the IEEE/CVF Conference on Computer Vision and Pattern Recognition}, pages 10491--10503, 2023.

\bibitem[\protect\citeauthoryear{Gu \bgroup \em et al.\egroup }{2021}]{SpatioTemporalInconsistency2021}
Zhihao Gu, Yang Chen, Taiping Yao, Shouhong Ding, Jilin Li, Feiyue Huang, and Lizhuang Ma.
\newblock Spatiotemporal inconsistency learning for deepfake video detection.
\newblock In {\em Proceedings of the 29th ACM international conference on multimedia}, pages 3473--3481, 2021.

\bibitem[\protect\citeauthoryear{Gu \bgroup \em et al.\egroup }{2022a}]{gu2022hierarchical}
Zhihao Gu, Taiping Yao, Yang Chen, Shouhong Ding, and Lizhuang Ma.
\newblock Hierarchical contrastive inconsistency learning for deepfake video detection.
\newblock In {\em European Conference on Computer Vision}, pages 596--613. Springer, 2022.

\bibitem[\protect\citeauthoryear{Gu \bgroup \em et al.\egroup }{2022b}]{gu2022region}
Zhihao Gu, Taiping Yao, C~Yang, Ran Yi, Shouhong Ding, and Lizhuang Ma.
\newblock Region-aware temporal inconsistency learning for deepfake video detection.
\newblock In {\em Proceedings of the 31th International Joint Conference on Artificial Intelligence}, volume~1, 2022.

\bibitem[\protect\citeauthoryear{Guarnera \bgroup \em et al.\egroup }{2020}]{guarnera2020deepfake}
Luca Guarnera, Oliver Giudice, and Sebastiano Battiato.
\newblock Deepfake detection by analyzing convolutional traces.
\newblock In {\em Proceedings of the IEEE/CVF conference on computer vision and pattern recognition workshops}, pages 666--667, 2020.

\bibitem[\protect\citeauthoryear{Hashmi \bgroup \em et al.\egroup }{2022}]{hashmi2022multimodal}
Ammarah Hashmi, Sahibzada~Adil Shahzad, Wasim Ahmad, Chia~Wen Lin, Yu~Tsao, and Hsin-Min Wang.
\newblock Multimodal forgery detection using ensemble learning.
\newblock In {\em 2022 Asia-Pacific Signal and Information Processing Association Annual Summit and Conference (APSIPA ASC)}, pages 1524--1532. IEEE, 2022.

\bibitem[\protect\citeauthoryear{Heo \bgroup \em et al.\egroup }{2023}]{heo2023deepfake}
Young-Jin Heo, Woon-Ha Yeo, and Byung-Gyu Kim.
\newblock Deepfake detection algorithm based on improved vision transformer.
\newblock {\em Applied Intelligence}, 53(7):7512--7527, 2023.

\bibitem[\protect\citeauthoryear{Ikram \bgroup \em et al.\egroup }{2023}]{ikram2023performance}
Sumaiya~Thaseen Ikram, Shourya Chambial, Dhruv Sood, et~al.
\newblock A performance enhancement of deepfake video detection through the use of a hybrid cnn deep learning model.
\newblock {\em International journal of electrical and computer engineering systems}, 14(2):169--178, 2023.

\bibitem[\protect\citeauthoryear{Jia \bgroup \em et al.\egroup }{2018}]{jia2018transfer}
Ye~Jia, Yu~Zhang, Ron Weiss, Quan Wang, Jonathan Shen, Fei Ren, Patrick Nguyen, Ruoming Pang, Ignacio Lopez~Moreno, Yonghui Wu, et~al.
\newblock Transfer learning from speaker verification to multispeaker text-to-speech synthesis.
\newblock {\em Advances in neural information processing systems}, 31, 2018.

\bibitem[\protect\citeauthoryear{Jung \bgroup \em et al.\egroup }{2023}]{jung2023safe}
Juho Jung, Chaewon Kang, Jeewoo Yoon, Simon~S Woo, and Jinyoung Han.
\newblock Safe: Sequential attentive face embedding with contrastive learning for deepfake video detection.
\newblock In {\em Proceedings of the 32nd ACM International Conference on Information and Knowledge Management}, pages 3993--3997, 2023.

\bibitem[\protect\citeauthoryear{Khalid \bgroup \em et al.\egroup }{2021a}]{khalid2021evaluation}
Hasam Khalid, Minha Kim, Shahroz Tariq, and Simon~S Woo.
\newblock Evaluation of an audio-video multimodal deepfake dataset using unimodal and multimodal detectors.
\newblock In {\em Proceedings of the 1st workshop on synthetic multimedia-audiovisual deepfake generation and detection}, pages 7--15, 2021.

\bibitem[\protect\citeauthoryear{Khalid \bgroup \em et al.\egroup }{2021b}]{khalid2021fakeavceleb}
Hasam Khalid, Shahroz Tariq, Minha Kim, and Simon~S Woo.
\newblock Fakeavceleb: A novel audio-video multimodal deepfake dataset.
\newblock {\em arXiv preprint arXiv:2108.05080}, 2021.

\bibitem[\protect\citeauthoryear{Khan and Dai}{2021}]{khan2021video}
Sohail~Ahmed Khan and Hang Dai.
\newblock Video transformer for deepfake detection with incremental learning.
\newblock In {\em Proceedings of the 29th ACM International Conference on Multimedia}, pages 1821--1828, 2021.

\bibitem[\protect\citeauthoryear{Korshunova \bgroup \em et al.\egroup }{2017}]{korshunova2017fast}
Iryna Korshunova, Wenzhe Shi, Joni Dambre, and Lucas Theis.
\newblock Fast face-swap using convolutional neural networks.
\newblock In {\em Proceedings of the IEEE International Conference on computer vision}, pages 3677--3685, 2017.

\bibitem[\protect\citeauthoryear{Kwon \bgroup \em et al.\egroup }{2021}]{kwon2021kodf}
Patrick Kwon, Jaeseong You, Gyuhyeon Nam, Sungwoo Park, and Gyeongsu Chae.
\newblock Kodf: A large-scale korean deepfake detection dataset.
\newblock In {\em Proceedings of the IEEE/CVF International Conference on Computer Vision}, pages 10744--10753, 2021.

\bibitem[\protect\citeauthoryear{Lewis \bgroup \em et al.\egroup }{2020}]{lewis2020deepfake}
John~K Lewis, Imad~Eddine Toubal, Helen Chen, Vishal Sandesera, Michael Lomnitz, Zigfried Hampel-Arias, Calyam Prasad, and Kannappan Palaniappan.
\newblock Deepfake video detection based on spatial, spectral, and temporal inconsistencies using multimodal deep learning.
\newblock In {\em 2020 IEEE Applied Imagery Pattern Recognition Workshop (AIPR)}, pages 1--9. IEEE, 2020.

\bibitem[\protect\citeauthoryear{Li \bgroup \em et al.\egroup }{2020}]{SharpMultiInstance2020}
Xiaodan Li, Yining Lang, Yuefeng Chen, Xiaofeng Mao, Yuan He, Shuhui Wang, Hui Xue, and Quan Lu.
\newblock Sharp multiple instance learning for deepfake video detection.
\newblock In {\em Proceedings of the 28th ACM international conference on multimedia}, pages 1864--1872, 2020.

\bibitem[\protect\citeauthoryear{Li \bgroup \em et al.\egroup }{2021a}]{li2021detecting}
Bo~Li, Jianghe Xu, Shuang Wu, Shouhong Ding, Jilin Li, and Feiyue Huang.
\newblock Detecting adversarial patch attacks through global-local consistency.
\newblock In {\em Proceedings of the 1st International Workshop on Adversarial Learning for Multimedia}, pages 35--41, 2021.

\bibitem[\protect\citeauthoryear{Li \bgroup \em et al.\egroup }{2021b}]{li2021deepfake}
Meng Li, Beibei Liu, Yongjian Hu, Liepiao Zhang, and Shiqi Wang.
\newblock Deepfake detection using robust spatial and temporal features from facial landmarks.
\newblock In {\em 2021 IEEE International Workshop on Biometrics and Forensics (IWBF)}, pages 1--6. IEEE, 2021.

\bibitem[\protect\citeauthoryear{Nguyen \bgroup \em et al.\egroup }{2022}]{SurveyDLforDF2022}
Thanh~Thi Nguyen, Quoc Viet~Hung Nguyen, Dung~Tien Nguyen, Duc~Thanh Nguyen, Thien Huynh-The, Saeid Nahavandi, Thanh~Tam Nguyen, Quoc-Viet Pham, and Cuong~M Nguyen.
\newblock Deep learning for deepfakes creation and detection: A survey.
\newblock {\em Computer Vision and Image Understanding}, 223:103525, 2022.

\bibitem[\protect\citeauthoryear{Raza and Malik}{2023}]{raza2023multimodaltrace}
Muhammad~Anas Raza and Khalid~Mahmood Malik.
\newblock Multimodaltrace: Deepfake detection using audiovisual representation learning.
\newblock In {\em Proceedings of the IEEE/CVF Conference on Computer Vision and Pattern Recognition}, pages 993--1000, 2023.

\bibitem[\protect\citeauthoryear{Seow \bgroup \em et al.\egroup }{2022}]{seow2022comprehensive}
Jia~Wen Seow, Mei~Kuan Lim, Raphael~CW Phan, and Joseph~K Liu.
\newblock A comprehensive overview of deepfake: Generation, detection, datasets, and opportunities.
\newblock {\em Neurocomputing}, 513:351--371, 2022.

\bibitem[\protect\citeauthoryear{Shahzad \bgroup \em et al.\egroup }{2022}]{shahzad2022lip}
Sahibzada~Adil Shahzad, Ammarah Hashmi, Sarwar Khan, Yan-Tsung Peng, Yu~Tsao, and Hsin-Min Wang.
\newblock Lip sync matters: A novel multimodal forgery detector.
\newblock In {\em 2022 Asia-Pacific Signal and Information Processing Association Annual Summit and Conference (APSIPA ASC)}, pages 1885--1892. IEEE, 2022.

\bibitem[\protect\citeauthoryear{Singh \bgroup \em et al.\egroup }{2020}]{singh2020using}
Simranjeet Singh, Rajneesh Sharma, and Alan~F Smeaton.
\newblock Using gans to synthesise minimum training data for deepfake generation.
\newblock {\em arXiv preprint arXiv:2011.05421}, 2020.

\bibitem[\protect\citeauthoryear{Tarasiou and Zafeiriou}{2020}]{tarasiou2020extracting}
Michail Tarasiou and Stefanos Zafeiriou.
\newblock Extracting deep local features to detect manipulated images of human faces.
\newblock In {\em 2020 IEEE international conference on image processing (ICIP)}, pages 1821--1825. IEEE, 2020.

\bibitem[\protect\citeauthoryear{Tolosana \bgroup \em et al.\egroup }{2020}]{tolosana2020deepfakes}
Ruben Tolosana, Ruben Vera-Rodriguez, Julian Fierrez, Aythami Morales, and Javier Ortega-Garcia.
\newblock Deepfakes and beyond: A survey of face manipulation and fake detection.
\newblock {\em Information Fusion}, 64:131--148, 2020.

\bibitem[\protect\citeauthoryear{Wang \bgroup \em et al.\egroup }{2022}]{wang2022m2tr}
Junke Wang, Zuxuan Wu, Wenhao Ouyang, Xintong Han, Jingjing Chen, Yu-Gang Jiang, and Ser-Nam Li.
\newblock M2tr: Multi-modal multi-scale transformers for deepfake detection.
\newblock In {\em Proceedings of the 2022 international conference on multimedia retrieval}, pages 615--623, 2022.

\bibitem[\protect\citeauthoryear{Yang \bgroup \em et al.\egroup }{2023}]{yang2023avoid}
Wenyuan Yang, Xiaoyu Zhou, Zhikai Chen, Bofei Guo, Zhongjie Ba, Zhihua Xia, Xiaochun Cao, and Kui Ren.
\newblock Avoid-df: Audio-visual joint learning for detecting deepfake.
\newblock {\em IEEE Transactions on Information Forensics and Security}, 18:2015--2029, 2023.

\bibitem[\protect\citeauthoryear{Yu \bgroup \em et al.\egroup }{2021}]{yu2021survey}
Peipeng Yu, Zhihua Xia, Jianwei Fei, and Yujiang Lu.
\newblock A survey on deepfake video detection.
\newblock {\em Iet Biometrics}, 10(6):607--624, 2021.

\bibitem[\protect\citeauthoryear{Yu \bgroup \em et al.\egroup }{2023}]{yu2023modality}
Cai Yu, Peng Chen, Jiahe Tian, Jin Liu, Jiao Dai, Xi~Wang, Yesheng Chai, and Jizhong Han.
\newblock Modality-agnostic audio-visual deepfake detection.
\newblock {\em arXiv preprint arXiv:2307.14491}, 2023.

\bibitem[\protect\citeauthoryear{Zhang \bgroup \em et al.\egroup }{2021}]{zhang2021detecting}
Daichi Zhang, Chenyu Li, Fanzhao Lin, Dan Zeng, and Shiming Ge.
\newblock Detecting deepfake videos with temporal dropout 3dcnn.
\newblock In {\em IJCAI}, pages 1288--1294, 2021.

\bibitem[\protect\citeauthoryear{Zhang}{2022}]{zhang2022deepfake}
Tao Zhang.
\newblock Deepfake generation and detection, a survey.
\newblock {\em Multimedia Tools and Applications}, 81(5):6259--6276, 2022.

\bibitem[\protect\citeauthoryear{Zhao \bgroup \em et al.\egroup }{2022}]{zhao2022self}
Hanqing Zhao, Wenbo Zhou, Dongdong Chen, Weiming Zhang, and Nenghai Yu.
\newblock Self-supervised transformer for deepfake detection.
\newblock {\em arXiv preprint arXiv:2203.01265}, 2022.

\bibitem[\protect\citeauthoryear{Zheng \bgroup \em et al.\egroup }{2021}]{zheng2021exploring}
Yinglin Zheng, Jianmin Bao, Dong Chen, Ming Zeng, and Fang Wen.
\newblock Exploring temporal coherence for more general video face forgery detection.
\newblock In {\em Proceedings of the IEEE/CVF international conference on computer vision}, pages 15044--15054, 2021.

\bibitem[\protect\citeauthoryear{Zhou and Lim}{2021}]{zhou2021joint}
Yipin Zhou and Ser-Nam Lim.
\newblock Joint audio-visual deepfake detection.
\newblock In {\em Proceedings of the IEEE/CVF International Conference on Computer Vision}, pages 14800--14809, 2021.

\end{thebibliography}
